\documentclass[letterpaper]{article} % DO NOT CHANGE THIS
\usepackage{aaai2026}  % DO NOT CHANGE THIS
\usepackage{times}  % DO NOT CHANGE THIS
\usepackage{helvet}  % DO NOT CHANGE THIS
\usepackage{courier}  % DO NOT CHANGE THIS
\usepackage[hyphens]{url}  % DO NOT CHANGE THIS
\usepackage{graphicx} % DO NOT CHANGE THIS
\usepackage{natbib}  % DO NOT CHANGE THIS AND DO NOT ADD ANY OPTIONS TO IT
\usepackage{caption} % DO NOT CHANGE THIS AND DO NOT ADD ANY OPTIONS TO IT
\usepackage{algorithm}
\usepackage{algorithmic}
\usepackage{amsmath} 
\usepackage{booktabs}
\usepackage{multirow}
\usepackage{amsmath}
\usepackage{cleveref} 
\usepackage{afterpage}
\usepackage{amssymb}
\usepackage{multirow}
\usepackage{booktabs}
\usepackage{pifont}
\usepackage{colortbl}

\usepackage{newfloat}
\usepackage{listings}
\DeclareCaptionStyle{ruled}{labelfont=normalfont,labelsep=colon,strut=off} % DO NOT CHANGE THIS
\floatstyle{ruled}
\newfloat{listing}{tb}{lst}{}
\floatname{listing}{Listing}
\title{Revisiting Lightweight Low-Light Image Enhancement: From a YUV Color
Space Perspective}

\author{
    Hailong Yan\textsuperscript{\rm 1,\rm 2}, Shice Liu\textsuperscript{\rm 2}, Xiangtao Zhang\textsuperscript{\rm 1}, \\
    Lujian Yao\textsuperscript{\rm 2}, Fengxiang Yang\textsuperscript{\rm 2}, Jinwei Chen\textsuperscript{\rm 2}, Bo Li\textsuperscript{\rm 2}\equalcontrib 
}
\affiliations{
    \textsuperscript{\rm 1}UESTC  ~ \textsuperscript{\rm 2}vivo Mobile Communication Co. Ltd\\

}

\usepackage{bibentry}
\begin{document}

\maketitle

\begin{figure*}[ht]
\centering
\includegraphics[width=1\linewidth]{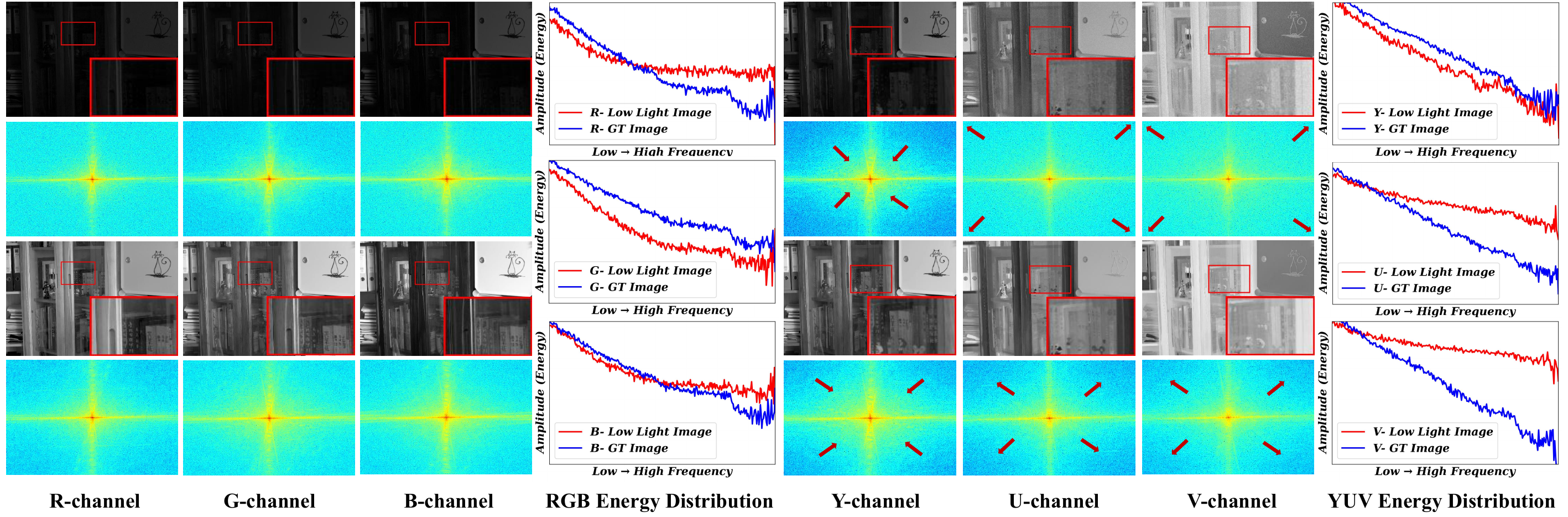} 
\caption{
Comparison of degradation patterns across RGB and YUV color spaces under low-light conditions. The first two rows show the low-light input images and their corresponding frequency spectrums (via Fourier Transform), while the last two rows show the ground truth (GT) images and their spectrums. Unlike the RGB channel-entangled degradation, YUV channels exhibit clearer disentangled: the Y channel mainly shows low-frequency luminance degradation, accompanied by potential high-frequency noise, while the U/V channels contain more prominent high-frequency chrominance noise.}
\label{fig1}
\vspace{-6pt}
\end{figure*}

\begin{abstract}
In the current era of mobile internet, Lightweight Low-Light Image Enhancement (L3IE) is critical for mobile devices, which faces a persistent trade-off between visual quality and model compactness. While recent methods employ disentangling strategies to simplify lightweight architectural design, such as Retinex theory and YUV color space transformations, their performance is fundamentally limited by overlooking channel-specific degradation patterns and cross-channel interactions. To address this gap, we perform a frequency-domain analysis that confirms the superiority of the YUV color space for L3IE. We identify a key insight: the Y channel primarily loses low-frequency content, while the UV channels are corrupted by high-frequency noise. Leveraging this finding, we propose a novel YUV-based paradigm that strategically restores channels using a Dual-Stream Global-Local Attention module for the Y channel, a Y-guided Local-Aware Frequency Attention module for the UV channels, and a Guided Interaction module for final feature fusion. Extensive experiments validate that our model establishes a new state-of-the-art on multiple benchmarks, delivering superior visual quality with a significantly lower parameter count.
\end{abstract}


\section{Introduction}

Lightweight Low-Light Image Enhancement (L3IE) aims to improve image quality under poor lighting conditions in real time on low-computing-power devices. With the increasing demand for mobile camera equipment such as drones and recorders, L3IE is becoming increasingly important.

Early L3IE methods, such as those relying on heuristic rules \cite{lee2013contrast}, often yielded limited enhancement quality. With the advent of deep learning, subsequent research efforts \cite{wei2018deep,fu2023learning} began to leverage the modeling capabilities of deep networks to directly recover brightness from low-exposure RGB images. However, these methods consistently faced a challenging trade-off: lightweight models tended to produce sub-optimal results, while high-performance models were too computationally intensive to be considered truly lightweight.

Recently, a paradigm shift towards disentanglement-based modular designs has dominated L3IE research. Retinex-based methods \cite{cai2023retinexformer} are a representative class, which disentangle RGB images into illumination and reflectance components for separate processing. A significant drawback, however, is that achieving effective disentanglement often necessitates deeper or multi-branch architectures, thereby increasing computational complexity.

An alternative line of work involves methods based on the YUV color space transformation. Inspired by the human visual system where rod cells perceive luminance and cone cells perceive color, these methods process the luminance (Y) and chrominance (UV) channels independently. Despite their promise, most YUV-based methods suffer from two principal limitations. First, they often apply homogeneous processing to all channels, and overlook perceptual differences across channels \cite{guo2023low,niu2025adaptive}. Second, other methods \cite{wang2024division,zhu2024ghost} optimize the luminance and chrominance channels in isolation, failing to establish effective cross-channel interaction, which hinders joint optimization and overall performance.

In order to guide the design of a more effective yet lightweight model, our work begins with an in-depth analysis of RGB and YUV degradation characteristics under low-light conditions, leveraging frequency-domain analysis. As illustrated in Fig. \ref{fig1}, the RGB channels exhibit highly similar degradation trends, indicating strong inter-channel entanglement. This complicates the simultaneous tasks of brightness enhancement and noise suppression, which presents a significant challenge for efficient network design. In contrast, the YUV channels offer an inherent advantage by naturally disentangling these degradations. Specifically, in low-light scenarios, the Y channel primarily suffers from a loss of low-frequency energy and a minor increase in high-frequency noise, which visually manifests as diminished brightness. Conversely, the UV channels are predominantly corrupted by substantial high-frequency noise, while their low-frequency components remain relatively stable.

Based on the aforementioned observations, we propose a novel YUV-based L3IE paradigm designed to systematically address low- and high-frequency degradations in low-light images. Our architecture is composed of three specialized modules. Firstly, a Dual-Stream Global-Local Attention (DSGLA) module is introduced to process the Y channel, which comprises a Dilated-Depth Self Attention (DDSA) to capture long-range dependencies and a Ghost-Gated Aggregation (GGA) to enhance local detail awareness. This dual-stream design makes it possible to effectively model the global low-frequency structures and the local high-frequency details. Secondly, to address the high-frequency noise dominating the UV channels, we introduce a Local-Aware Frequency Attention (LAFA) module. The LAFA incorporates guidance from the enhanced Y-channel features, and leverages a Frequency Channel Attention (FCA) mechanism on Fourier features to improve denoising. Thirdly, a Guided Interaction (GI) module orchestrates the fusion of Y and UV features via cross-channel and frequency-aware interactions, yielding a coherently enhanced image.

In summary, our main contributions are:
\begin{itemize}
\item Through a comprehensive frequency-domain analysis of low-light degradation, we demonstrate that addressing the L3IE task within the YUV color space facilitates the design of more compact and effective models.
\item We introduce a novel YUV-based L3IE paradigm featuring channel-specific processing for Y and UV, culminating in an effective fusion. Our 30K-parameter model sets a new benchmark for efficiency, requiring fewer parameters than even the most lightweight existing models.
\item Extensive experiments show that our strategic design yields a model that achieves highly competitive performance against SOTA methods on all benchmarks.
\end{itemize}

\section{Related Works}
\label{sec:formatting}

\textbf{End-to-end RGB-based methods.} These methods leverage the fitting capabilities of neural networks to directly process RGB images and produce brightness-enhanced results. \cite{lore2017llnet} first adopted an autoencoder for joint denoising and brightening, followed by CNN-based methods~\cite{zhang2019kindling, guo2020zero, yan2025mobileie} that improved detail restoration through end-to-end or self-supervised learning. Considering CNNs are limited in modeling long-range dependencies, Transformer-based approaches~\cite{wang2023ultra, liu2024efficient} address this by leveraging global self-attention. \cite{xu2022snr} incorporated signal-to-noise priors for adaptive enhancement. In addition, diffusion-based methods~\cite{wang2024zero, lv2024fourier, hou2023global} have shown promising results in image quality and detail restoration. However, either transformers or diffusion models often suffer from high computational cost. In conclusion, end-to-end RGB-based methods face difficulties in effectively handling the complex interplay between brightness and noise.

\textbf{Retinex-based L3IE.} As a prominent disentanglement strategy for L3IE, Retinex-based L3IE \cite{zhao2021retinexdip,cai2023retinexformer,fu2023learning,liu2024efficient, yan2024igdnet, liu2024ntire} disentangled an RGB image into reflectance and illumination for individual processing. These approaches theoretically allows robust structure preservation while modifying brightness. In practice, however, its efficacy is limited by two fundamental challenges. First, the Retinex decomposition itself is an ill-posed problem, since its simple priors like illumination smoothness frequently result in artifacts and unrealistic lighting. Second, stabilizing this ill-posed decomposition often demands complex, parameter-rich networks, directly contradicting the core objective of lightweight design.

\begin{figure*}[ht]
\centering
\includegraphics[width=1\linewidth]{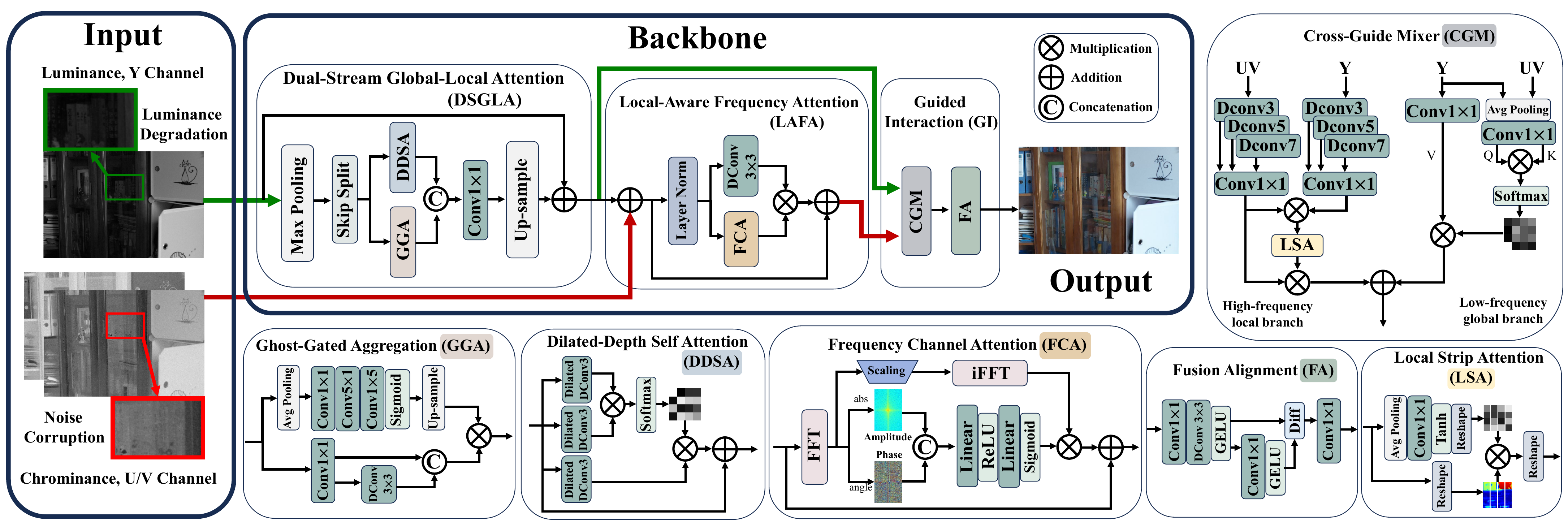} 
\caption{The overview of the proposed method. The Y channel is enhanced via DSGLA, while the UV channels are refined by LAFA with guidance from Y features. The GI module fuses luminance and chrominance features for consistent enhancement.}
\label{fig2}
% \vspace{-8pt}
\end{figure*}

\textbf{YUV-based L3IE.} The YUV color space offers a compelling alternative to Retinex, as its explicit separation of luminance and chrominance bypasses the ill-posed decomposition problem. This allows for a targeted brightness adjustment while preserving color fidelity \cite{guo2023low,zhang2021better,niu2025adaptive}. However, the promise of this approach is frequently unrealized as current methods are hampered by two critical flaws. First, they often employ channel-agnostic designs, applying uniform operations to Y and UV channels despite their distinct degradation patterns. Second, many recent works \cite{wang2024division, zhu2024ghost, wang2024extracting, brateanu2025lyt} treat the luminance and chrominance pathways as isolated pipelines, lacking the sophisticated interaction necessary for coherent restoration. This oversight often leads to a suboptimal outcome, where enhancements in brightness come at the cost of color fidelity, or vice versa.

\section{Proposed Method}
\subsection{Overall Pipeline}

As shown in Fig.~\ref{fig2}, the proposed framework consists of three modules: Dual-Stream Global-Local Attention \textbf{(DSGLA)}, Local-Aware Frequency Attention \textbf{(LAFA)}, and Guided Interaction \textbf{(GI)}, corresponding to the luminance branch, chrominance branch, and cross-channel fusion. A low-light RGB image $I_{\mathrm{low}}^{\mathrm{RGB}} \in \mathbb{R}^{3 \times H \times W}$ is first converted to YUV color space. The \textbf{DSGLA} enhances the Y channel by addressing low-frequency luminance degradation via global-local attention, producing $I_{\text{enh}}^Y$. The \textbf{LAFA} takes $I_{\text{enh}}^Y$ as guidance and the original UV input $I_{\text{low}}^{\text{UV}}$ to suppress high-frequency noise via frequency-aware attention, producing $I_{\text{enh}}^{\text{UV}}$. Finally, the \textbf{GI} fuses $I_{\text{enh}}^Y$ and $I_{\text{enh}}^{\text{UV}}$ via cross-channel interaction, producing the final output $I_{\text{enh}}^{\text{RGB}}$.

\subsection{DSGLA: Dual-Stream Global-Local Attention}

Given that luminance degradation in the Y channel is generally global and low-frequency, while residual noise is local and high-frequency, the DSGLA module is designed to separately capture structural and detail cues. A max pooling operation is first applied to the Y-channel feature to emphasize dominant low-frequency structures and suppress irrelevant high-frequency noise, facilitating global context modeling and reducing computation in later stages.

Unlike \cite{brateanu2025lyt} that processes the Y channel via a single pathway, DSGLA separates global structures and local details via a \textit{channel-splitting} design. Specifically, the input feature $X \in \mathbb{R}^{C \times H \times W}$ is split along the channel dimension into two parts. $X_{\text{Gl}}$ is used for modeling global structures and $X_{\text{Lo}}$ focuses on refining local details.
\begin{equation}
X_{\text{Gl}},\ X_{\text{Lo}} = \text{Split}(X),\quad X_{\text{Gl}},\ X_{\text{Lo}} \in \mathbb{R}^{\frac{C}{2} \times H \times W}.
\end{equation}

\noindent\textbf{A. Global branch: Dilated-Depth Self Attention (DDSA)}

To efficiently capture long-range dependencies in luminance features, the global branch adopts a self-attention mechanism enhanced by dilation and channel grouping. Specifically, given the input $X_{\text{Gl}}$, we first apply a $1 \times 1$ pointwise convolution followed by a dilated depthwise convolution (DDConv) to generate the Q, K, and V features:
\begin{equation}
Q, K, V = \text{DDConv}_{3}(\text{Conv}_{1}(X_{\text{Gl}})).
\end{equation}

The dilated convolution expands the receptive field without increasing parameters, while depthwise and grouped operations reduce computational overhead, making the design both global-aware and efficient. The Q, K, and V tensors are reshaped and fed into a multi-head self-attention block:
\begin{equation}
\text{Attention}(Q, K, V) = \text{Softmax}({QK^\top}/{\sqrt{d}}) V,
\end{equation}
where $d$ is the dimensionality per head. The output of DDSA is referred to as $F_{Gl}$. As self-attention excels at modeling global low-frequency dependencies, this branch is well-suited for capturing the large-scale luminance degradation in the Y channel, achieving enhanced structural sensitivity with reduced computational cost.

\noindent\textbf{B. Local branch: Ghost-Gated Aggregation (GGA)}

The Ghost-style structure efficiently reuses and enriches features, making it suitable for capturing fine-grained visual details with low computational cost \cite{han2020ghostnet}. Motivated by this, we design a \textbf{GGA} module that integrates the Ghost-style structure with a spatial gating mechanism to enhance local high-frequency representations.

While self-attention architectures are effective in modeling global dependencies and low-frequency semantics, convolutional operations excel at capturing high-frequency details (edges and textures), through local receptive fields \cite{li2025exploring}. Leveraging this property, the Ghost-style convolution further extends conventional convolutions by generating additional feature maps with fewer parameters, thus reducing redundancy and enriching feature diversity.

\begin{figure}[ht]
\centering
\includegraphics[width=1\linewidth]{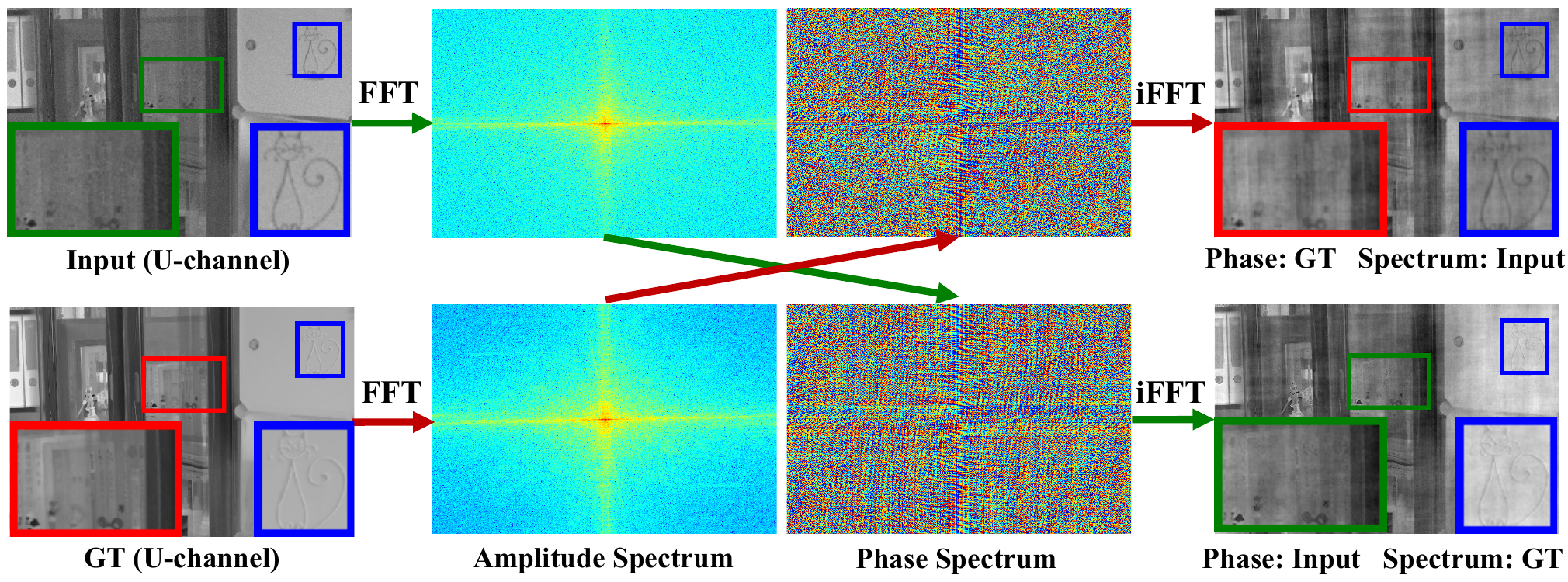} 
\caption{Visualization of spectrum swapping experiments.}
\label{fig3}
\vspace{-4pt}
\end{figure}

First, the input feature $X_{\text{Lo}} \in \mathbb{R}^{C \times H \times W}$ is processed through two parallel branches to generate ghost features:
\begin{equation}
\begin{array}{l}
F_{\mathrm{base}} = \mathrm{Conv}_{1}(X_{\text{Lo}}), \\
F_{\mathrm{ghost}} = \mathrm{Concat}(F_{\mathrm{base}},\ \mathrm{DConv}_{3}(F_{\mathrm{base}})).
\end{array}
\end{equation}

Here, $F_{\mathrm{base}}$ represents the primary feature map, while the depthwise convolution ($\mathrm{DConv}$) generates complementary features with fewer parameters. The concatenated output $F_{\mathrm{ghost}}$ encodes both efficiency and feature diversity.

Then, a spatial modulation map $M$ is computed from a low-resolution shortcut pathway:
\begin{equation}
M = \mathrm{Sigmoid}(\left(\text{Up}\left(f_{\text{short}}(\text{AvgPool}(X_{\text{Lo}})\right)\right)),
\end{equation}
where $f_{\text{short}}(\cdot)$ consists of a series of grouped asymmetric convolutions (e.g., $1 \times 5$, $5 \times 1$), $\text{Up}(\cdot)$ denotes upsample.

Finally, the output is modulated as: $F_{Lo} = F_{\text{ghost}} \odot M$.

The outputs of $F_{Gl}$ and $F_{Lo}$ from the DDSA and GGA modules are concatenated and passed through a $1 \times 1$ convolution for channel integration. This fusion combines global low-frequency structures and local high-frequency details, improving the representation for subsequent stages.

\begin{table*}[htbp]
  \centering
  \resizebox{\textwidth}{!}{
    \begin{tabular}{ccc|cc|ccc|ccc|ccc}
    \toprule[2pt]
    \multirow{2}{*}{Method} & \multirow{2}{*}{Venue} & \multirow{2}{*}{Category} & \multicolumn{2}{c|}{Complexity} & \multicolumn{3}{c|}{LOLv1} & \multicolumn{3}{c|}{LOLv2-Real} & \multicolumn{3}{c}{LOLv2-Syn} \\
    \cmidrule{4-5} \cmidrule{6-8} \cmidrule{9-11} \cmidrule{12-14}
    & & & Params$\downarrow$ & FLOPs$\downarrow$ & PSNR$\uparrow$ & SSIM$\uparrow$ & LPIPS$\downarrow$ & PSNR$\uparrow$ & SSIM$\uparrow$ & LPIPS$\downarrow$ & PSNR$\uparrow$ & SSIM$\uparrow$ & LPIPS$\downarrow$ \\
    \hline
    Zero-DCE \cite{guo2020zero} & CVPR'20 & RGB & 0.075 & 4.83 & 21.880 & 0.640 & 0.335 & 16.059 & 0.580 & 0.313 & 17.712 & 0.815 & 0.169  \\
    SNR-Net \cite{xu2022snr}& CVPR'22 & RGB & 4.01 & 26.35 & 26.716 & \underline{0.851} & 0.152 & 27.349 & 0.871 & 0.151 & 27.830 & \underline{0.942} & \textbf{0.051} \\
    LPF \cite{dang2023you} & ICIP'23 & RGB & 0.082 & 4.83 & 25.659 & 0.848 & 0.133 & 28.015 & 0.879 & 0.166 & 28.159 & 0.939 & 0.062  \\
    LLFormer \cite{wang2023ultra} & AAAI'23 & RGB & 24.55 & 22.52 & 26.106 & 0.830 & 0.166 & 29.307 & 0.866 & 0.141 & 27.574 & 0.936 & 0.065  \\
    UHDFour \cite{UHDFourICLR2023} & ICLR'23 & RGB & 15.90 & 57.42 & 26.308 & 0.836 & 0.143 & 27.472 & 0.878 & 0.162 & 27.087 & 0.912 & 0.078  \\
    FourierDiff \cite{lv2024fourier} & CVPR'24 & RGB & 547.55 & - & 22.142 & 0.660 & 0.265 & 22.551 & 0.679 & 0.259 & 16.790 & 0.735 & 0.245  \\
    LIME \cite{lime} & TIP'16 & Retinex & - & - & 18.947 & 0.448 & 0.408 & 18.495 & 0.469 & 0.402 & 18.391 & 0.773 & 0.218  \\
    RetinexFormer \cite{cai2023retinexformer}& ICCV'23 & Retinex & 1.53 & 15.85 & \underline{27.140} & 0.850 & \textbf{0.129} & 27.690 & 0.857 & 0.166 & \underline{29.024} & 0.939 & 0.055 \\
    PairLIE \cite{fu2023learning} & CVPR'23 & Retinex & 0.33 & 20.81 & 23.526 & 0.755 & 0.248 & 24.026 & 0.803 & 0.227 & 21.681 & 0.820 & 0.224  \\
    CCNet \cite{jin2024colorization} & TNNLS'24 & LAB & 3.78 & - & 26.665 & 0.842 & 0.156 & 27.963 & 0.878 & 0.149 & 27.982 & 0.935 & 0.058   \\
    CTNet \cite{xie2025ctnet} & PR'25 & HSV & 5.62 & 62.14 & 27.042 & 0.849 & 0.135 & 28.296 & 0.899 & 0.134 & 28.968 & 0.938 & 0.064  \\
    QuadPrior \cite{wang2024zero} & CVPR'24 & K-M & 1252.75 & 1103.20 & 22.849 & 0.800 & 0.201 & 23.633 & 0.829 & 0.197 & 19.131 & 0.809 & 0.224  \\
    Bread \cite{guo2023low} & IJCV'23 & YUV & 2.02 & 19.85 & 25.299 & 0.847 & 0.155 & 26.916 & 0.883 & 0.152 & 19.379 & 0.810 & 0.243  \\
    Ghillie \cite{zhu2024ghost} & TCSVT'24 & YUV & 3.05 & 199.36 & 25.182 & 0.840 & 0.155 & 28.584 & 0.880 & 0.126 & 24.627 & 0.905 & 0.104  \\
    END \cite{wang2024extracting} & TCSVT'24  & YUV & 7.91 & 270.44 & 26.364 & 0.850 & 0.133 & \underline{30.117} & \underline{0.895} & \underline{0.110} & 28.738 & 0.939 & \underline{0.053} \\
    LYTNet \cite{brateanu2025lyt} & SPL'25 & YUV & \underline{0.045} & \underline{1.70} & 26.581 & 0.835 & 0.133 & 28.342 & 0.877 & 0.127 & 26.671 & 0.927 & 0.081 \\
    \rowcolor{gray!20}
    Ours &  & YUV & \textbf{0.030} & \textbf{1.45} & \textbf{27.160} & \textbf{0.853} & \underline{0.131} & \textbf{31.181} & \textbf{0.896} & \textbf{0.104} & \textbf{29.081} & \textbf{0.943} & \textbf{0.051} \\
    \toprule[2pt]
    \end{tabular}
  }
  \caption{Performance comparison of different L3IE models. The best results are in \textbf{bold}, and the second-best are \underline{underlined}. The units for Param and FLOPs are M and G. Models above the line are large-scale, those below are lightweight.}
  \label{tab:1}
\end{table*}

\subsection{LAFA: Local-Aware Frequency Attention}

In the UV channels, where degradation is dominated by high-frequency noise, frequency-domain modeling facilitates a more explicit disentanglement of signal and noise components. This not only promotes better distinguishing valid chrominance structures from interference, but also enhances its ability to suppress fine-grained degradation.

However, some frequency-domain methods \cite{qiao2023learning, zhang2025prnet} focus exclusively on the amplitude spectrum, potentially overlooking the structural cues embedded in the phase. To assess the role of the phase, we conduct a spectrum replacement experiment, where only the amplitude is replaced by that of a GT image. The reconstruction preserves structural details consistent with the image providing the phase, as shown in the blue boxes (Fig.~\ref{fig3}). This indicates that the phase spectrum aids structural alignment.

Therefore, we propose the LAFA to enhance the UV channels via luminance-guided frequency-domain modulation. Unlike previous methods that process UV features independently, LAFA incorporates the refined Y-channel features $I_{\text{enh}}^Y$ from the DSGLA as structural priors. These features provide stable high-frequency guidance to help distinguish signals from high-frequency noise in chrominance.

Concretely, the LAFA takes the chrominance feature $I_{\text{low}}^{UV}$ and the enhanced luminance feature $I_{\text{enh}}^Y$ from DSGLA as input. We first perform addition between $I_{\text{low}}^{UV}$ and $I_{\text{enh}}^Y$ to obtain a fused representation, which is transformed into the frequency domain:~$X_A,\ X_P = \mathrm{FFT}(I_{\text{low}}^{UV} + I_{\text{enh}}^Y)$, where $X_A$ and $X_P$ denote the amplitude and phase, respectively. We apply global average pooling to both components and concatenate them to form the spectral feature $f$. 
\begin{equation}
f = \mathrm{Concat}\left(\mathrm{AvgPool}(X_A),\ \mathrm{AvgPool}(X_P)\right).
\end{equation}

And then, the features $f$ are passed through a two-layer MLP to produce channel-wise attention weights $W$. 
\begin{equation}
W = \mathrm{Sigmoid}(\left( \mathrm{FC}_2\left( \mathrm{ReLU}(\mathrm{FC}_1(f)) \right) \right)).
\end{equation}

In parallel, a learnable modulation mask $\hat{M} \in \mathbb{R}^{1 \times C \times 1 \times 1}$ is equipped to $X_A$, $X_P$ to scale the spectrum. The spatial feature $X'$ is recovered via the inverse FFT.

%, where the modulated frequency representation is given by $\hat{\mathcal{F}} = \mathrm{FFT}(I_{\text{low}}^{UV} + I_{\text{enh}}^Y) \cdot \hat{M}$. : $X' = \mathrm{iFFT}(\hat{\mathcal{F}})$

To further enhance local spatial consistency and suppress fine-grained chrominance noise, we apply a 3$\times$3 depthwise convolution to the same fused input. The final output is obtained by element-wise multiplication of both branches:
\begin{equation}
X_{\text{LAFA}} = X' \cdot W \cdot \mathrm{DConv}_{3}(I_{\text{low}}^{UV} + I_{\text{enh}}^Y) + (I_{\text{low}}^{UV} + I_{\text{enh}}^Y).
\end{equation}

\subsection{GI: Guided Interaction}

For the purpose of achieving coordinated enhancement of structure and color, different from most YUV-based methods which process luminance and chrominance separately, we propose the Guided Interaction (GI) to explore the complementary potential of luminance and chrominance features. The GI consists of two sub-modules: the Cross-Guide Mixer (CGM) and the Fusion Alignment (FA) module.

The CGM consists of two branches. The global branch employs Transformer-style cross attention to refine low-frequency structures in the Y channel, guided by the relatively clean low-frequency cues from the UV features. This cross-channel interaction facilitates the restoration of degraded luminance. The local branch employs Local Strip Attention (LSA), which operates on horizontal strips rather than dense spatial correlations. Benefiting from the high-frequency details from the Y channel, LSA efficiently enhances fine-grained UV textures, ensuring that chrominance information is consistent with luminance details.

\begin{figure*}[t]
\centering
\includegraphics[width=1\linewidth]{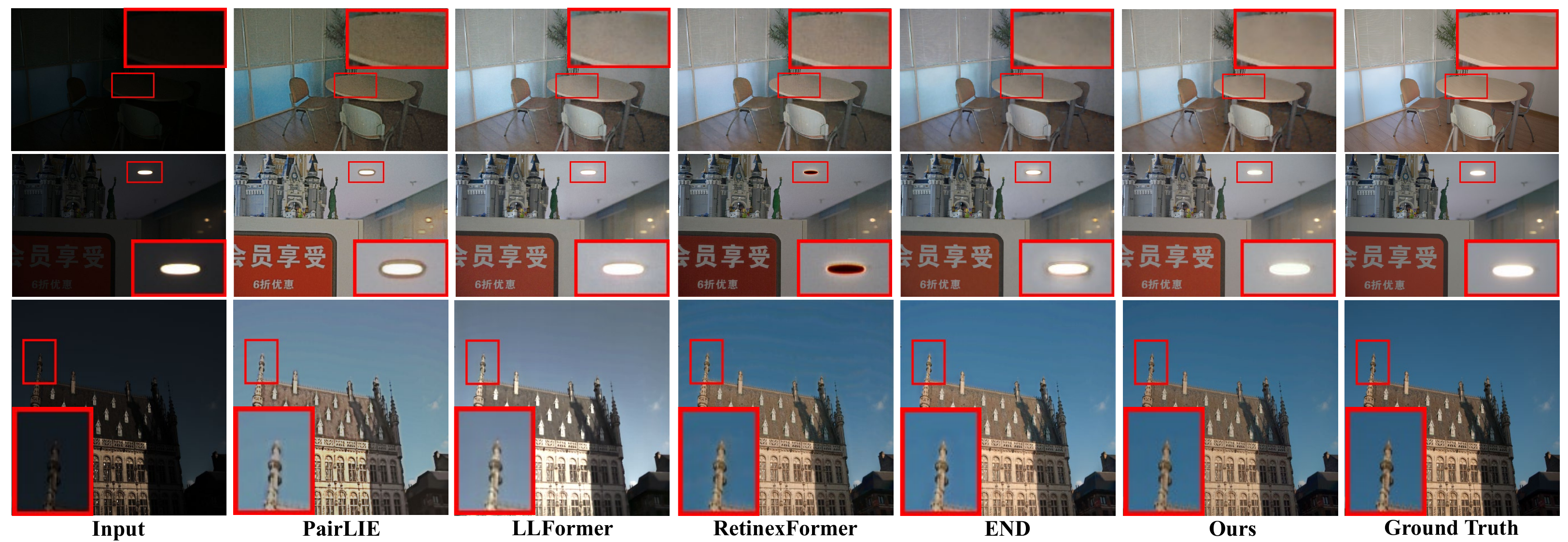} 
\caption{Visual comparison of L3IE methods on the LOL dataset. Red boxes highlight noise suppression.}
\label{fig4}
\vspace{-6pt}
\end{figure*}

To guide UV refinement, we extract high-frequency components from both $I^Y_{\text{enh}}$ and $I^{UV}_{\text{enh}}$ using multi-scale depthwise convolutions, denoted as ($\text{F}_{K:3}^{\text{Y}}, \text{F}_{K:5}^{\text{Y}}, \text{F}_{K:7}^{\text{Y}}$) and ($\text{F}_{K:3}^{\text{UV}}, \text{F}_{K:5}^{\text{UV}}, \text{F}_{K:7}^{\text{UV}}$), respectively. These are concatenated as:
\begin{equation}
\begin{array}{l}
\text{F}_Y^{\text{high}} = \text{Conv}_1(\text{Concat}(\text{F}_{K:3}^{\text{Y}}, \text{F}_{K:5}^{\text{Y}}, \text{F}_{K:7}^{\text{Y}})), \\
\text{F}_{UV}^{\text{high}} = \text{Conv}_1(\text{Concat}(\text{F}_{K:3}^{\text{UV}}, \text{F}_{K:5}^{\text{UV}}, \text{F}_{K:7}^{\text{UV}})).
\end{array}
\end{equation}

These are then fused by element-wise multiplication to form: $\text{F} = \text{F}_Y^{\text{high}} \cdot \text{F}_{UV}^{\text{high}}$, which is subsequently processed by the LSA module. LSA generates a channel-wise strip attention map by applying a $1 \times 1$ convolution followed by $\mathrm{Tanh}$ activation on pooled features, allowing attention to focus on horizontal strips and enhance fine-grained UV textures:
\begin{equation}
\text{F}_{l}^{H} = \text{F}_{UV}^{\text{high}} \cdot \sum\nolimits_{i=1}^{k} \text{P}_i \cdot \mathrm{Tanh}(\mathrm{Conv}_{1}(\mathrm{AvgPool}(\text{F}))_i,
\end{equation}

\begin{figure}[htbp]
\centering
\includegraphics[width=1\linewidth]{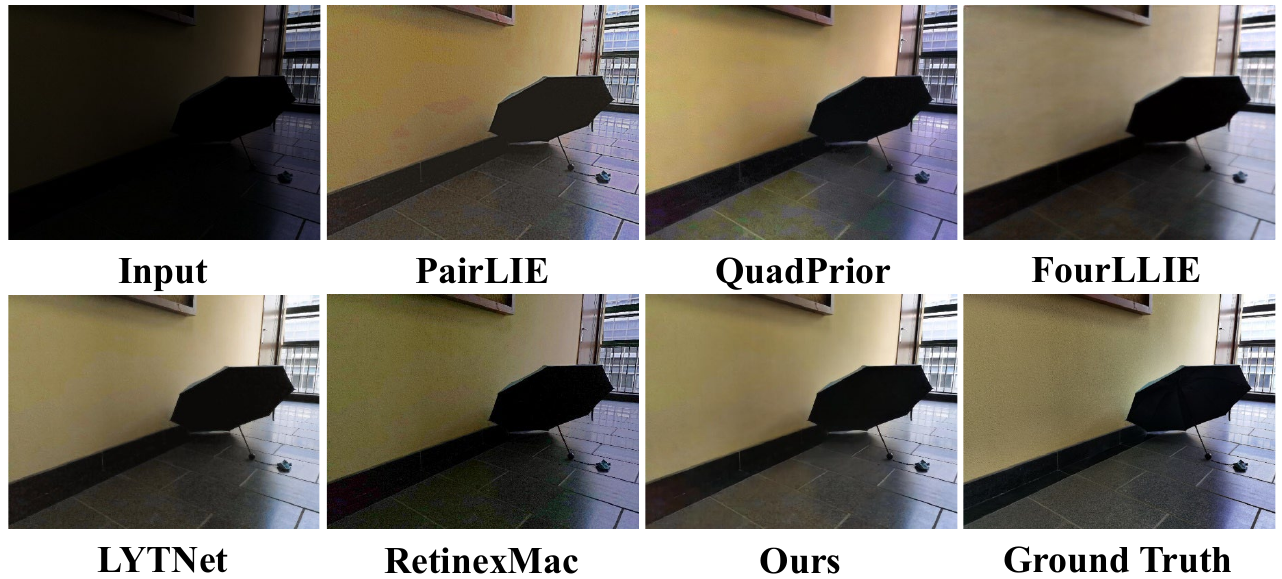} 
\caption{Visual comparison on LSRW.}
\label{fig5}
\vspace{-8pt}
\end{figure}

where $P_i$ denotes the unfolded horizontal patch at position $i$, and $k$ is the kernel width. The output of CGM ($\text{F}_{YUV}$) is the sum of the global output $\text{F}_{g}^{L}$ and the local output $\text{F}_{l}^{H}$:
\begin{equation}
\text{F}_{YUV} = \text{F}_{g}^{L} + \text{F}_{l}^{H}.
\end{equation}

To ensure the consistency of luminance and chrominance information in the output, FA combines channel aggregation with offset removal to optimize the coordination of YUV channel features. Specifically, FA removes redundancy by calculating the difference between features and activated features, avoiding inconsistencies between luminance and chrominance channels. This process can be expressed as:
\begin{equation}
\begin{array}{l}
\bar{\mathcal{F}} = \text{GELU}(\text{DConv}_3(\text{Conv}_1(\text{F}_{YUV}))), \\
\text{O} = \bar{\mathcal{F}} + \gamma \cdot \left( \bar{\mathcal{F}} - \text{GELU}(\text{Conv}_1(\bar{\mathcal{F}})) \right).
\end{array}
\end{equation}

where \(\bar{\mathcal{F}}\) is the refined feature, and \(\gamma\) is a learnable scaling factor. The final output \(\text{O}\) is the adjusted feature map that ensures the alignment of luminance and chrominance.

\begin{figure*}[t]
\centering
\includegraphics[width=1\linewidth]{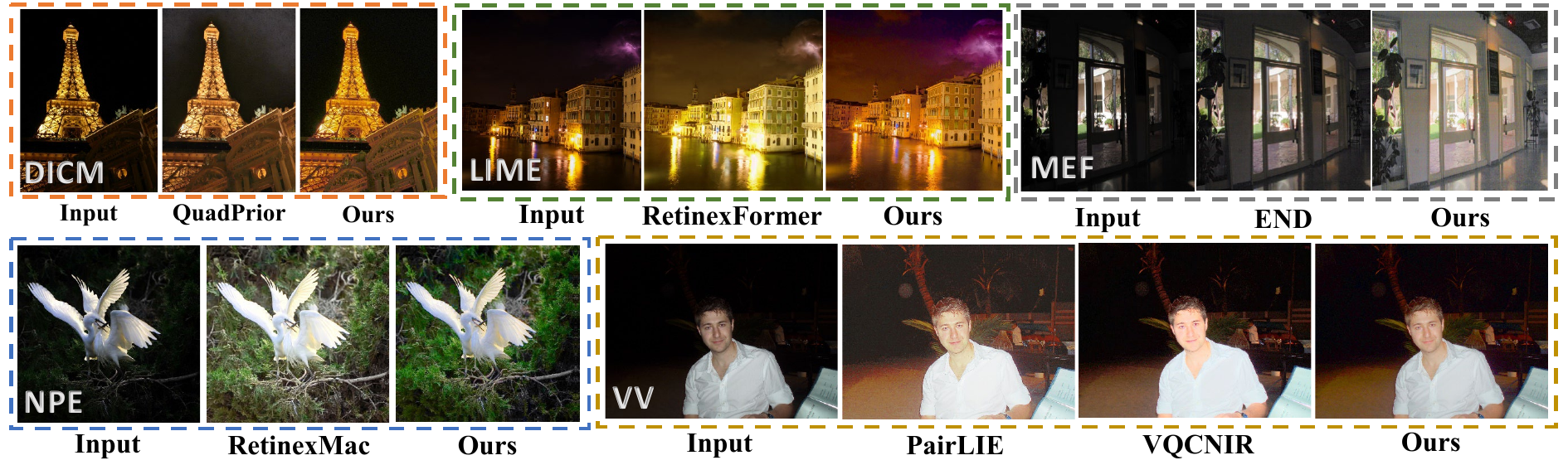} 
\caption{Visual comparison on unpaired datasets. Following RetinexFormer, one image per dataset is selected for comparison.}
\label{fig6}
\vspace{-8pt}
\end{figure*}

\section{Experiments}
\subsection{Experimental Settings}
\textbf{Implementation Details.} We implement our model on PyTorch and train it with the Adam optimizer ($\beta_{1}=0.9$, $\beta_{2}=0.999$) on an NVIDIA RTX~3090 GPU for 5,000 epochs, with an initial learning rate of $2 \times 10^{-4}$. During training, images are randomly cropped to $256 \times 256$ with a batch size of 1, and the network is optimized under the joint supervision of Smooth L1 and PSNR losses.

\begin{table}[htbp]
\centering
\resizebox{\linewidth}{!}{
    \begin{tabular}{c|cc|cc|cc}
    \toprule[2pt]
    \multirow{2}{*}{Method} & \multicolumn{2}{c|}{Complexity} & \multicolumn{2}{c|}{LSRW (Nikon)} & \multicolumn{2}{c}{LSRW (Huawei)} \\
    \cmidrule{2-7}
    & Params$\downarrow$ & FLOPs$\downarrow$ & PSNR$\uparrow$ & SSIM$\uparrow$ & PSNR$\uparrow$ & SSIM$\uparrow$ \\
    \hline
    Zero-DCE \cite{guo2020zero} & 0.075 & 4.83 & 15.04 & 0.420 & 16.40 & 0.476 \\
    EnGAN \cite{jiang2021enlightengan} & 114.35 & 61.01 & 14.63 & 0.398 & 17.46 & 0.498 \\
    SNR-Net \cite{xu2022snr} & 4.01 & 26.35 & 16.63 & 0.505 & 20.40 & 0.617 \\
    PairLIE \cite{fu2023learning} & 0.33 & 20.81 & 15.52 & 0.435 & 18.99 & 0.563 \\
    FourLLIE \cite{wang2023fourllie} & 1.49 & 16.29 & \underline{17.82} & 0.504 & 21.11 & 0.626 \\
    RetinexFormer \cite{cai2023retinexformer} & 1.53 & 15.85 & 17.64 & 0.508 & 21.23 & 0.631 \\
    WaveMamba \cite{zou2024wave} & 1.52 & 15.97 & 17.34 & 0.519 & 21.19 & 0.639 \\
    QuadPrior \cite{wang2024zero} & 1252.75 & 1103.20 & 14.84 & 0.487 & 18.31 & 0.600 \\
    RetinexMac \cite{liu2024efficient} & 2.29 & 29.59 & 16.66 & 0.461 & 19.07 & 0.596 \\
    DMFourLLIE \cite{zhang2024dmfourllie} & 0.97 & 8.36 & 17.04 & \textbf{0.529} & 21.47 & 0.633 \\
    LYTNet \cite{brateanu2025lyt} & \underline{0.045} & \underline{1.70} & 17.69 & 0.513 & 21.06 & 0.627 \\
    DarkIR \cite{feijoo2025darkir} & 3.32 & 7.25 & 16.12 & 0.507 & 20.78 & 0.625 \\
    CWNet \cite{zhang2025cwnet} & 1.23 & 11.30 & - & - & \underline{21.51} & \underline{0.640} \\
    \rowcolor{gray!20}
    Ours & \textbf{0.03} & \textbf{1.45} & \textbf{17.99} & \underline{0.525} & \textbf{21.52} & \textbf{0.642} \\
    \toprule[2pt]
    \end{tabular}
  }
  \caption{Performance comparison on the LSRW dataset. Best values are in \textbf{bold}, second-best are \underline{underlined}.}
  \label{tab:2}
  \vspace{-8pt}
\end{table}

\textbf{Datasets}. To evaluate the effectiveness of our method, we conduct experiments on both paired and unpaired datasets. The paired datasets include LOLv1 \cite{wei2018deep}, LOLv2 \cite{yang2020fidelity}, and LSRW \cite{hai2023r2rnet}. For unpaired evaluation, we utilize the DICM \cite{lee2013contrast}, LIME \cite{lime}, MEF \cite{ma2015perceptual}, NPE \cite{wang2013naturalness}, and VV.

\textbf{Metrics}. Consistent with prior work, we evaluate our model using six key metrics to ensure a comprehensive assessment. We assess restoration quality on paired data using PSNR for pixel-wise fidelity, SSIM for structural similarity, and the perceptual metric LPIPS \cite{zhang2018unreasonable}, which captures perceptual differences more effectively than pixel-based metrics. Model efficiency is measured via Parameters (Params) and FLOPs, which indicate the model's memory usage and computational cost, respectively. Additionally, the non-reference metric NIQE \cite{mittal2012making} is employed specifically for quality assessment on unpaired datasets.

\subsection{Quantitative and Visual Comparisons}

\textbf{Results on Pair Datasets}. As shown in Tables \ref{tab:1} and \ref{tab:2}, our method performs excellently on both LOL and LSRW. On LOL, we achieve the highest average PSNR of 29.141 dB and the second-lowest LPIPS of 0.095, while on LSRW, we obtain PSNRs of 17.99 dB and 21.52 dB, and SSIMs of 0.525 and 0.642, respectively. Using only 0.03M parameters and 1.45 GFLOPs, our method balances performance and efficiency. Compared to larger models like RetinexFormer (1.53M parameters, 15.85 GFLOPs), it reduces computational cost while maintaining quality. Compared to smaller models like LYTNet (0.045M parameters, 1.70 GFLOPs), our method outperforms them in both datasets, showing a clear advantage in performance and efficiency.

\begin{table}[htbp]
\centering
\resizebox{\linewidth}{!}{
    \begin{tabular}{c|cc|ccccc|c}
    \toprule[2pt]
    Method & Params$\downarrow$ & FLOPs$\downarrow$ & DICM & LIME & MEF & NPE & VV & Average$\downarrow$ \\
    \hline
    Zero-DCE \cite{guo2020zero} & \underline{0.075} & 4.83 & 3.535 & 4.271 & 3.680 & \underline{4.048} & 3.927 & 3.892 \\
    RetinexDIP \cite{zhao2021retinexdip} & 0.71 & - & 3.383 & \underline{3.741} & 3.669 & 4.947 & 3.975 & 3.943 \\
    SNR-Net \cite{xu2022snr} & 4.01 & 26.35 & 3.586 & 5.844 & 4.060 & 4.700 & 5.589 & 4.756 \\
    Bread \cite{guo2023low} & 2.02 & 19.85 & 3.737 & 4.528 & 4.229 & 4.293 & 3.596 & 4.077 \\
    PairLIE \cite{fu2023learning} & 0.33 & 20.81 & 4.120 & 4.516 & 4.176 & 4.358 & 3.662 & 4.116 \\
    Retinexformer \cite{cai2023retinexformer} & 1.53 & 15.85 & 3.651 & 4.276 & 4.230 & 4.381 & 3.981 & 4.104 \\
    GSAD \cite{hou2023global} & 17.43 & - & 3.902 & 4.547 & 4.339 & 4.736 & 4.051 & 4.315 \\
    SWANet \cite{he2023low} & 6.86 & 55.40 & \underline{3.169} & 4.786 & 4.100 & 4.121 & 3.335 & 3.902 \\
    VQCNIR \cite{zou2024vqcnir} & 45.93 & 162.46 & 3.961 & 4.103 & 4.680 & 4.258 & 3.346 & 4.070 \\
    RetinexMac \cite{liu2024efficient} & 2.29 & 29.59 & \textbf{3.064} & 4.191 & \underline{3.613} & \textbf{3.777} & \underline{3.201} & \underline{3.569} \\
    \rowcolor{gray!20}
    Ours & \textbf{0.030} & \textbf{1.45} & 3.623 & \textbf{3.594} & \textbf{3.334} & 4.071 & \textbf{3.126} & \textbf{3.550} \\
    \bottomrule[2pt]
    \end{tabular}
}
\caption{NIQE comparison on five unpair datasets. Best values are in \textbf{bold}, second-best are \underline{underlined}.}
\label{tab:3}
\vspace{-6pt}
\end{table}

Fig.~\ref{fig4} and \ref{fig5} show the visual comparisons on the LOL \cite{wei2018deep,yang2020fidelity} and LSRW \cite{hai2023r2rnet} datasets. Our method enhances brightness while preserving dark region details, achieving more balanced and natural results. Unlike methods like PairLIE \cite{fu2023learning} and QuadPrior \cite{wang2024zero}, which struggle with uniform brightness, our approach avoids overexposure and effectively improves low-light areas. It also suppresses noise in dark regions better than methods like LLFormer \cite{wang2023ultra}, producing cleaner, less grainy images. Additionally, our method maintains natural color balance, whereas other methods, such as RetinexMac \cite{liu2024efficient}, introduce color shifts, leading to more realistic and visually pleasing results.

\textbf{Results on Unpair Datasets}. As shown in Table~\ref{tab:3}, our method achieves an average NIQE score of 3.550 across five unpaired datasets, outperforming RetinexMac and VQCNIR. It achieves the lowest NIQE scores on DICM (3.535) and MEF (3.334), indicating superior image quality.

In Fig.~\ref{fig6}, our method enhances brightness, preserves fine details, and effectively reduces noise, especially in low-light areas. Compared to QuadPrior, it also maintains more natural color, resulting in clearer and more realistic images.

\begin{figure}[!ht]
\centering
\vspace{-2pt}
\includegraphics[width=1\linewidth]{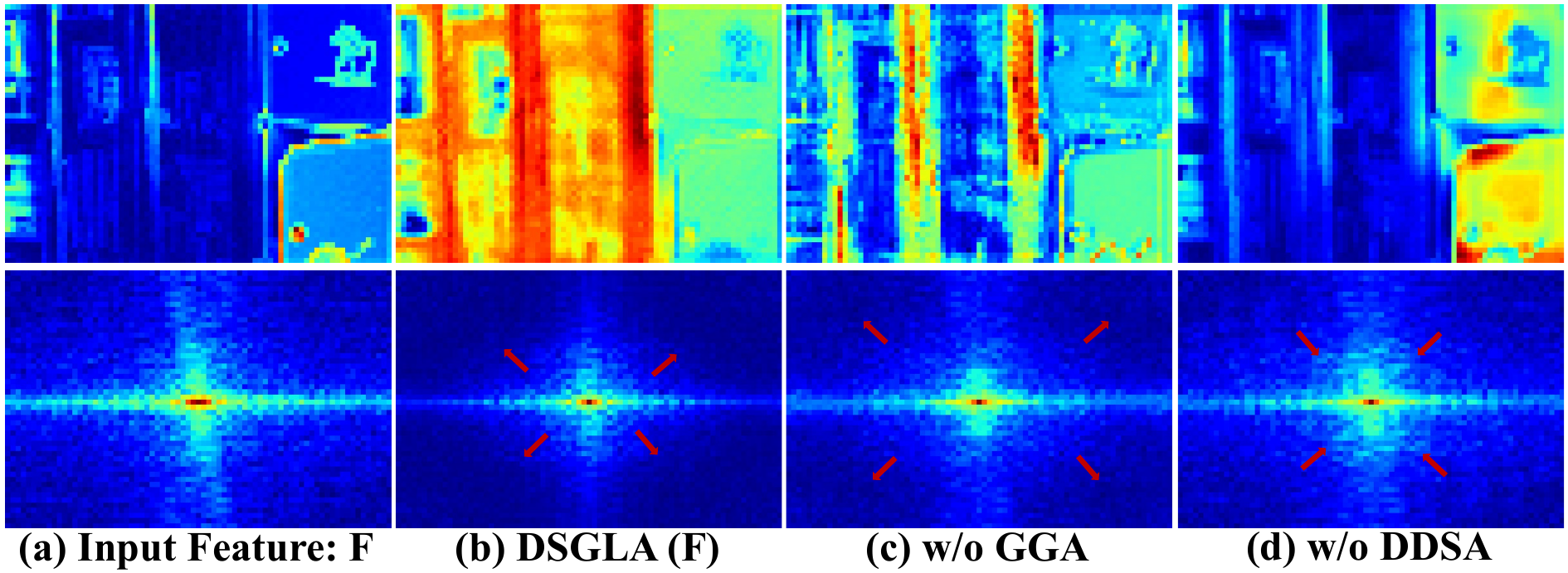} 
\caption{Feature Spectrum Visualization of DSGLA.}
\label{fig8}
\vspace{-4pt}
\end{figure}

\begin{table}[htb]
\centering
\begin{minipage}[t]{0.45\linewidth}
\centering
\scalebox{0.55}{
\begin{tabular}{c|ccc>{\columncolor{gray!20}}c}
\toprule[2pt]
Color & R-GB & G-RB & B-GR & Y-UV \\
\midrule
PSNR$\uparrow$  & 26.65 & \underline{26.69} & 26.35 & \textbf{27.16} \\
SSIM$\uparrow$  & 0.841 & \underline{0.847} & 0.842 & \textbf{0.853} \\
LPIPS$\downarrow$ & 0.152 & \underline{0.141} & 0.148 & \textbf{0.131} \\
\bottomrule[2pt]
\end{tabular}
}
\caption{Ablation study on different color space.}
\label{tab4}
\end{minipage}
\hfill
\begin{minipage}[t]{0.54\linewidth}
\centering
\scalebox{0.55}{
\begin{tabular}{c|ccc>{\columncolor{gray!20}}cc}
\toprule[2pt]
Ratios & 1 & 2 & 4 & \textbf{8} & 16\\
\midrule
PSNR$\uparrow$  & 26.53 & 26.82 & \underline{26.86} & \textbf{27.16} & 26.70 \\
SSIM$\uparrow$  & 0.845 & 0.838 & \underline{0.845} & \textbf{0.853} & 0.848 \\
LPIPS$\downarrow$ & 0.143 & 0.155 & \underline{0.142} & \textbf{0.131} & 0.135 \\
\bottomrule[2pt]
\end{tabular}
}
\caption{Ablation study on downsampling ratios.}
\label{tab5}
\end{minipage}
\end{table}

\subsection{Ablation Studies and Analyses}

\textbf{Replacing YUV with RGB.} We conducted a critical ablation study to empirically prove the superiority of the YUV color space over RGB for L3IE using our proposed architecture. As shown in Table~\ref{tab4}, we evaluated several RGB input configurations against our standard YUV input. The results are decisive: the YUV-based model significantly outperforms all RGB counterparts on all metrics. The G-RB configuration emerged as the best-performing RGB variant, a result consistent with the spectral similarity between the G and Y channels (Fig.~\ref{fig1}). This experiment provides compelling evidence that the performance gains of our model stem directly from the strategic advantages of channel-specific processing in the YUV space.

\begin{table}[h]
\centering
\resizebox{\linewidth}{!}{
\begin{tabular}{c|ccccc}
\toprule[2pt]
Method  & PSNR$\uparrow$ & SSIM$\uparrow$ & LPIPS$\downarrow$ & LOE$\downarrow$ & MAE$\downarrow$ \\
\hline
w/o DSGLA &  26.502 & 0.831 & 0.163 & 0.269 & 0.092 \\
w/o DDSA  & 26.643 & 0.845 & 0.138 & \underline{0.255} & 0.086\\
w/o GGA & 26.767 & \underline{0.849} & 0.134 & 0.262 & \underline{0.084}\\
HFERB \cite{li2023feature} & 26.725 & 0.849 & \underline{0.133} & 0.260 & 0.090\\
A-MSA \cite{wang2023ultra} & 26.555 & 0.842 & 0.139 & 0.262 & 0.098 \\
DHSA \cite{sun2024restoring} & \underline{26.908} & 0.846 & 0.134 & 0.258 & 0.094\\
\rowcolor{gray!20}
Ours & \textbf{27.160} & \textbf{0.853} & \textbf{0.131} & \textbf{0.254} & \textbf{0.081} \\
\bottomrule[2pt]
\end{tabular}
}
\caption{Quantitative Results of DSGLA Ablation Study.}
\label{tab:DSGLA}
\vspace{-6pt}
\end{table}

\textbf{Analysis of DSGLA.} To verify the effectiveness of DSGLA, we conduct ablation experiments on its submodules, with results shown in Table \ref{tab:DSGLA} and Fig.~\ref{fig8}.

Removing DSGLA causes a clear performance drop. Further, removing DDSA (PSNR:-0.517dB) weakens global structure modeling, while removing GGA (PSNR:-0.393dB) degrades local texture restoration. Our method also outperforms recent attention designs like DHSA.

We also analyze the impact of downsampling ratios (Table \ref{tab5}). The best performance is achieved with an 8×. Since the Y degradation mainly affects low-frequency luminance across the image, capturing the overall brightness distribution is crucial. MaxPooling allows for effective spatial compression while preserving essential structures and suppressing noise in low-contrast regions. A small ratio (2 or 4) fails to capture global luminance trends, while a large ratio (16) loses too much spatial detail, impairing texture recovery.

\begin{figure}[!ht]
\centering
\includegraphics[width=1\linewidth]{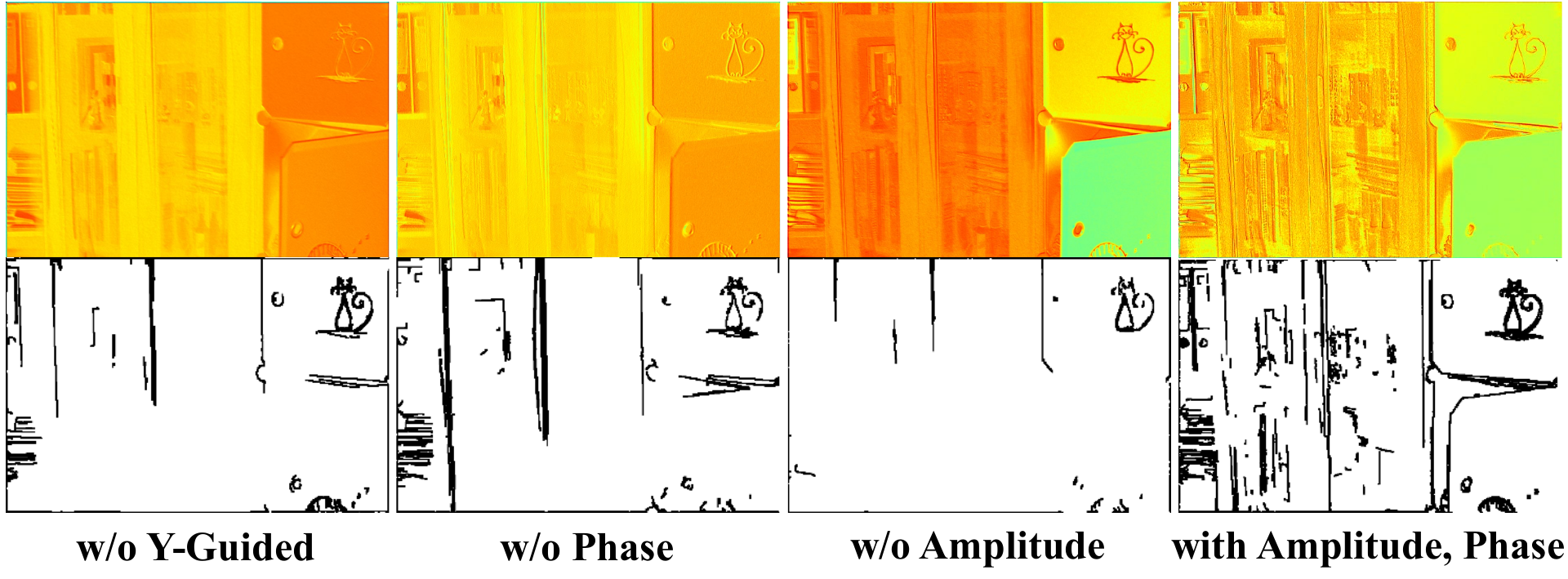} 
\caption{Visual of LAFA Effects on Feature Representation. Each block shows the LAFA output feature and the Canny edge of (output-input) feature difference (bottom-right).}
\label{fig9}
\vspace{-3pt}
\end{figure}

In Fig.\ref{fig8}(b), DSGLA effectively enhances high-frequency details and suppresses low-frequency degradation, aligning with the Y channel’s dominant low-frequency luminance interference. Without GGA, high-frequency texture responses are notably weakened; without DDSA, low-frequency information is insufficiently modeled.

\begin{table}[htbp]
\centering
\resizebox{\linewidth}{!}{
\begin{tabular}{c|cccc}
\toprule[2pt]
Method & PSNR$\uparrow$ & SSIM$\uparrow$ & LPIPS$\downarrow$ & LOE$\downarrow$ \\
\hline
w/o LAFA & 26.551 & 0.833 & 0.175 & 0.263 \\
w/o Phase & 26.748 & 0.841 & 0.149 & 0.263 \\
w/o Amplitude & 26.624 & 0.838 & 0.158 & \underline{0.255} \\
w/o Y-Guided & 26.426 & 0.838 & 0.155 & 0.261 \\
MCSF \cite{cui2023image} & 26.872 & \underline{0.850} & \underline{0.135} & 0.259 \\
EB \cite{feijoo2025darkir} & \underline{26.926} & \textbf{0.853} & 0.136 & \underline{0.255} \\
\rowcolor{gray!20}
Ours & \textbf{27.160} & \textbf{0.853} & \textbf{0.131} & \textbf{0.254} \\
\bottomrule[2pt]
\end{tabular}
}
\caption{Quantitative Results of LAFA Ablation Study.}
\label{tab:LAFA}
\vspace{-6pt}
\end{table}

\textbf{Analysis of LAFA.} We perform ablation studies to evaluate LAFA. As shown in Table~\ref{tab:LAFA}, removing amplitude weakens noise suppression, while omitting phase or Y-guided reduces structural consistency and detail alignment.

Fig.~\ref{fig9} presents the visual comparisons. Without amplitude, fine textures are lost and noise remains in flat regions. Without phase, edge structures become blurry and disconnected. Removing Y-guidance causes inconsistencies in structure across channels. In contrast, using all components results in sharper, more coherent edge representations.

\begin{table}[htbp]
\centering
\resizebox{0.95\linewidth}{!}{
\begin{tabular}{c|ccccc}
\toprule[2pt]
Method & PSNR$\uparrow$ & SSIM$\uparrow$ & LPIPS$\downarrow$ & LOE$\downarrow$ & MAE$\downarrow$\\
\hline
Concatenation & 26.070 & 0.814 & 0.191 & 0.268 & 0.092\\
Sum & 25.997 & 0.834 & 0.158 & 0.260 & 0.093\\
w/o FA & 26.467 & 0.839 & 0.142 & 0.259 & 0.095 \\
HFB \cite{li2025exploring} & \underline{26.725} & 0.850 & \underline{0.133} & 0.260 & 0.086\\
LCA \cite{yan2025hvi} & 26.627 & \underline{0.851} & 0.137 & \underline{0.255} & 0.089 \\
\rowcolor{gray!20}
Ours & \textbf{27.160} & \textbf{0.853} & \textbf{0.131} & \textbf{0.254} & \textbf{0.081}\\
\bottomrule[2pt]
\end{tabular}}
\caption{Quantitative Results of GI Ablation Study.}
\label{tab:gi}
\vspace{-6pt}
\end{table}

\textbf{Analysis of GI.} To evaluate the GI, we compare it with conventional fusion strategies (concat, sum) and recent fusion designs. As shown in Table~\ref{tab:gi}, naive fusion methods perform poorly due to their inability to fully leverage the complementary roles of the Y and UV channels. In contrast, GI achieves consistently better scores, highlighting its advantage in modeling cross-channel dependencies.

\begin{figure}[!ht]
\centering
\includegraphics[width=1\linewidth]{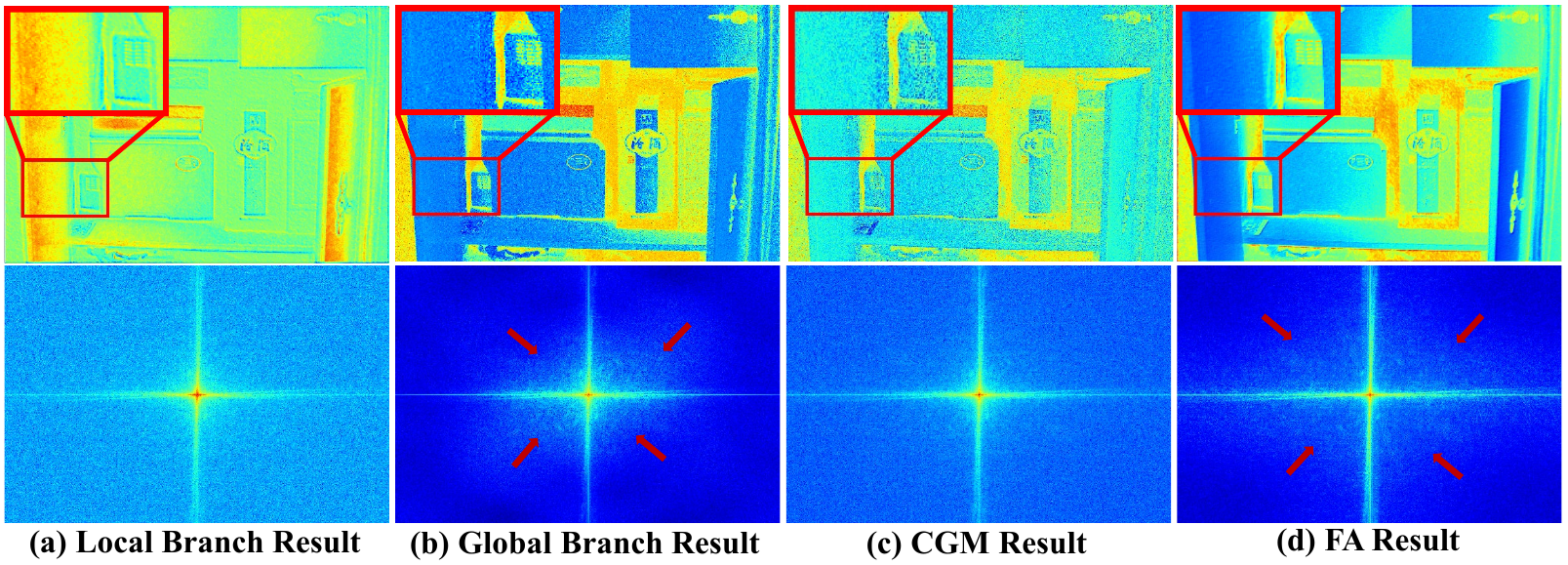} 
\caption{Feature Spectrum Visualization of GI.}
\label{fig10}
\vspace{-6pt}
\end{figure}

We further analyze the GI via frequency-domain visualizations, as shown in Fig.~\ref{fig10}. In the CGM, the local branch uses Y features to guide high-frequency enhancement in the UV channels, while the global branch leverages UV cues to refine low-frequency structures in the Y channel. This bidirectional guidance aligns structural and chromatic features across channels. The subsequent FA module suppresses redundant high-frequency noise and strengthens dominant low-frequency components, leading to sharper structural contours and more consistent color transitions.

\textbf{Latency Performance Evaluation.} Table~\ref{tab:speed} shows latency results on both GPU and CPU platforms. Our model achieves a latency of \textbf{6.5 ms} on GPU and \textbf{124.1 ms} on CPU, outperforming RetinexMac (32.4/739.3 ms) and LYTNet (7.9/165.8 ms). With only 30K parameters and 1.45 GFLOPs, it ensures efficient real-time processing on both high-performance and resource-constrained devices.

In contrast, other models with higher latency and larger parameters perform poorly in such environments. Our method offers a significant advantage in low-latency, efficient processing across different hardware platforms.

\begin{table}[htbp]
\centering
\resizebox{\linewidth}{!}{
\begin{tabular}{c|cccc>{\columncolor{gray!20}}c}
\toprule[2pt]
Method & RetinexMac & END & RetinexFormer & LYTNet & Ours\\
\hline
Latency (ms) & 32.4/739.3 & 92.7/- & 18.9/382.5 & \underline{7.9/165.8} & \textbf{6.5/124.1} \\
Params (M)& 2.29 & 7.91 & 1.53 & \underline{0.05} & \textbf{0.03} \\
\bottomrule[2pt]
\end{tabular}}
\caption{Latency comparison on both GPU/CPU platforms.}
\label{tab:speed}
\vspace{-4pt}
\end{table}

\textbf{Limitations.}
Our work validates a YUV-based framework, but its optimality relative to other color spaces remains an open question. We believe exploring alternative color representations is a promising avenue for future L3IE research to achieve even greater performance.

\section{Conclusion}
This paper revisits the lightweight low-light image enhancement (L3IE) task from the color space perspective and introduces a novel YUV-based approach. The proposed framework concludes three modules: DSGLA capturing global-local structures in the Y channel, LAFA suppressing high-frequency noise in the UV channel via illumination-guided frequency-aware modulation, and GI enabling structure-aware fusion between Y and UV. With only 30K parameters, our proposed framework not only outperforms other SOTA L3IE methods in low-light enhancement but also achieves an excellent balance between visual quality and model compactness across multiple benchmarks.

\bibliography{aaai2026}

\end{document}